\documentclass{tTMB2e}
\ifdefined\pdfinfoomitdate\pdfinfoomitdate=1\fi
\ifdefined\pdfsuppressptexinfo\pdfsuppressptexinfo=15\fi
\ifdefined\pdftrailerid\pdftrailerid{}\fi

\usepackage[T1]{fontenc}
\usepackage{lmodern}
\usepackage[utf8]{inputenc}
\usepackage[british]{babel}
\usepackage{booktabs}
\usepackage{tabularx}
\usepackage{array}
\usepackage{multirow}
\usepackage{xurl}
\usepackage{xcolor}
\newcommand{\StudyRMSE}{7.41}
\newcommand{\StudyMAPE}{31.24}
\newcommand{\StudyCorrelation}{0.987}
\newcommand{\StudyInitialMean}{1.0}
\newcommand{\StudyRMSEImprovement}{40.1}
\usepackage[hidelinks]{hyperref}

\hypersetup{
  pdfcreator={},
  pdfproducer={},
  pdftitle={Online Estimation of Dynamic Origin-Destination Matrices Using Reinforcement Learning with Link-Flow Propagation Guidance},
  pdfauthor={Donggyu Min and Dong-Kyu Kim}
}

\title{Online Estimation of Dynamic Origin-Destination Matrices Using Reinforcement Learning with Link-Flow Propagation Guidance}
\received{}
\author{%
Donggyu Min and Dong-Kyu Kim$^{\ast}$%
\thanks{$^\ast$Corresponding author. Email: \href{mailto:dongkyukim@snu.ac.kr}{dongkyukim@snu.ac.kr}}\\
{\em Seoul National University, Seoul, 08826, South Korea}\\
\texttt{dgmin@snu.ac.kr}\quad \texttt{dongkyukim@snu.ac.kr}
}

\begin{document}
\maketitle

\begin{abstract}
Dynamic origin--destination (OD) matrix estimation calibrates time-dependent input demand for simulations to reproduce observed link flows. Reinforcement learning is well suited to online estimation because a trained policy estimates demand in a single evaluation, but varying target flows complicate learning from aggregate rewards. We propose reinforcement learning with link-flow propagation guidance (LFPG-RL), combining simulated vehicle propagation records with downstream errors to guide individual OD components. Using 250 weekday link-flow trajectories from a Melbourne arterial network, LFPG-RL achieves a mean test root mean squared error (RMSE) of \StudyRMSE{} vehicles per 15-minute interval, mean absolute percentage error of \StudyMAPE\%, and Pearson correlation of \StudyCorrelation{}. Comparisons with optimisation, filtering and reinforcement learning without guidance show a \StudyRMSEImprovement\% RMSE reduction over the strongest baseline, gradient descent with propagation guidance. LFPG-RL enables rapid demand updates that keep traffic simulations consistent with observed conditions, supporting the evaluation of traffic management strategies.

\end{abstract}


\section{Introduction}

Traffic simulations used to evaluate route guidance, signal control, and other management strategies require calibration to observed conditions \citep{ref1, ref2}. Dynamic origin--destination (OD) matrix estimation (DODE) provides time-dependent demand inputs that reproduce observed link flows \citep{ref3, ref4, ref5}.

Estimating these demand inputs relies on dynamic network loading (DNL), whose nonlinear, delayed, and often stochastic response makes calibration computationally demanding \citep{ref6, ref7, ref8}. This is particularly challenging in online DODE, where estimation must be completed within short decision intervals \citep{ref33}. Existing methods improve computational efficiency \citep{ref5, ref8, ref9, ref10, ref11, ref12}, but many require repeated updates or simulator evaluations as new observations arrive.

Reinforcement learning (RL) offers an alternative by learning a demand estimation policy through interaction with a traffic simulator \citep{ref13}. The trained policy maps current link-flow observations and the simulated network state to an OD demand vector in a single policy evaluation.

For RL-based online DODE, the central challenge is to learn a policy that generalises across target link-flow trajectories because a deployed policy must respond to new observations without being retrained for each day. An OD component affects several downstream links over successive intervals, and its contribution to the error changes with the target trajectory. Aggregate rewards obscure these contributions, making it difficult to learn how individual demand estimates should change.

To address this difficulty, we propose LFPG-RL, which incorporates link-flow propagation guidance (LFPG) into RL. A DNL rollout records how vehicles from each OD pair and departure interval contribute to downstream link flows. LFPG combines these records with flow errors to guide individual OD components through advantage shaping in a component-wise update based on proximal policy optimisation (PPO). The actor update retains the global advantage from scalar rewards and adds the OD-specific guidance.

The contributions are as follows:
\begin{enumerate}
    \item \textbf{\textit{Extending RL-based DODE to online estimation}}. We train a policy across varying link-flow trajectories to estimate current OD demand from available observations and the simulated network state, avoiding repeated online optimisation.
    \item \textbf{\textit{Introducing OD-specific guidance for policy learning}}. To address credit assignment under varying targets, we construct LFPG from simulated vehicle propagation records and downstream flow errors and incorporate it into component-wise PPO updates.
    \item \textbf{\textit{Improving estimation accuracy and online efficiency}}. Tests with Melbourne detector data show a \StudyRMSEImprovement\% reduction in link-flow RMSE relative to LFPG-GD, the strongest baseline. Median online processing takes 0.012 s per interval, including DNL, compared with approximately 27--30 s for iterative methods.
\end{enumerate}

\section{Related work}

\subsection{Optimisation-based Online DODE}

Optimisation-based DODE searches for demand inputs that minimise the discrepancy between simulated and observed link flows. Each candidate demand is evaluated through dynamic traffic assignment (DTA) or DNL, accounting for route choice, congestion, and time-dependent propagation. These evaluations are computationally expensive, and analytical gradients are often unavailable. Methods developed to reduce the cost of offline calibration are therefore relevant to online estimation under limited per-interval budgets.

Simultaneous perturbation stochastic approximation (SPSA) estimates gradients through demand perturbations. Weighted SPSA (W-SPSA) uses spatial and temporal correlations to reduce gradient noise \citep{ref11}, c-SPSA perturbs homogeneous clusters of OD entries \citep{ref22}, and PC-SPSA reduces the search dimension using principal component analysis (PCA) \citep{ref23}. Other approaches use metamodels to reduce expensive simulation evaluations \citep{ref8} or PCA to estimate demand when reliable historical estimates are unavailable \citep{ref24}. Assignment-free DODE reduces the number of variables and avoids the classical bilevel assignment loop \citep{ref12}. Bayesian optimisation and hybrid macro-micro formulations have also been explored \citep{ref25, ref26}. These approaches reduce calibration cost but still require iterative search. For surrogate-based methods, approximation errors can also affect calibration quality \citep{ref27}.

\subsection{State-space and Filtering-based Online DODE}

Offline DODE estimates demand over a complete observation horizon. Online DODE instead updates demand sequentially as observations arrive, which has motivated recursive state-space methods.

Ashok and Ben-Akiva formulated real-time OD matrix estimation and prediction using deviations from historical demand as the state vector \citep{ref5}. Their later work incorporated the stochastic mapping from OD demand to path and link flows \citep{ref16}. Bierlaire and Crittin introduced an LSQR algorithm for efficient estimation \citep{ref10}, while Zhou and Mahmassani decomposed demand into regular patterns, structural deviations, and random fluctuations \citep{ref17}.

PCA reduces the dimension of demand variation, enabling reduced-space extended Kalman filtering (EKF) for real-time DTA calibration \citep{ref18, ref19}. Castiglione et al. combined PCA with the local ensemble transform Kalman filter (LETKF) in an assignment-matrix-free online framework \citep{ref20}. Zhang et al. improved EKF-based calibration using state augmentation and graph-colouring-based finite differences \citep{ref21}.

\subsection{Reinforcement Learning for Online DODE}

Optimisation and filtering methods update demand as observations arrive, but repeated evaluation of the demand--flow relationship can be costly within short estimation intervals. RL offers a promising alternative by learning a demand estimation policy before deployment. Min et al. formulated DODE as a Markov decision process (MDP) and trained a policy for offline calibration of microscopic traffic simulations \citep{ref13}. Their policy was fitted to a given target trajectory, leaving its extension to varying online observations unresolved.

For online DODE, the policy must estimate demand for new target trajectories while accounting for delayed effects of earlier demand. Training across trajectories changes both the link-flow errors and the OD components contributing to them, making a common estimation rule difficult to learn from aggregate rewards alone. Dynamic assignment information links OD inputs to downstream flows \citep{ref14}. This motivates LFPG-RL, which uses realised propagation records to construct OD-specific learning signals for policy updates.

\section{Methods}

\subsection{Key Concept}
\begin{figure}[htbp]
  \centering
  \includegraphics[width=1.0\linewidth]{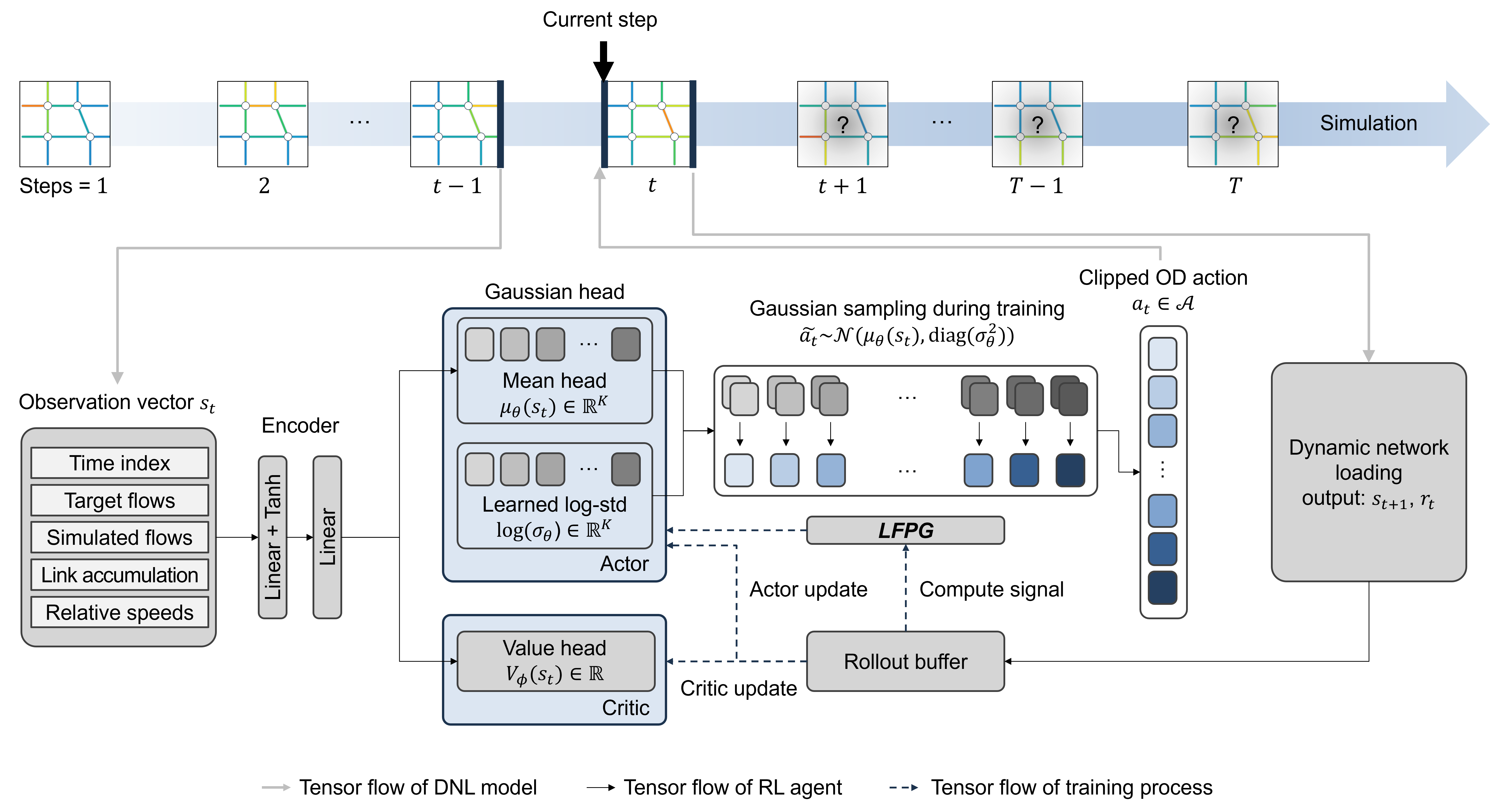}
  \caption{LFPG-RL workflow for sequential OD demand estimation and policy training. The policy receives current observations and the simulated network state, while completed DNL rollouts provide LFPG for actor updates.}
  \label{fig1}
\end{figure}

We formulate online DODE over $T$ decision steps. At step $t$, the estimator selects current OD demand using the available link-flow observation and the network state resulting from earlier decisions. The objective is to minimise cumulative link-flow error through sequential demand estimates, without revising earlier decisions or using future observations. Figure~\ref{fig1} shows the resulting LFPG-RL workflow. The policy estimates demand, DNL propagates it through the network, and the completed training rollout provides propagation records and flow errors for the policy update.

\subsection{Problem Statement}

\subsubsection{Notation}
Table~\ref{tab:t1} defines the notation.

\begin{table}[!t]
  \centering
  \caption{Notation}
  \label{tab:t1}
  \footnotesize
  \setlength{\tabcolsep}{2pt}
  \renewcommand{\arraystretch}{0.92}

  \begin{tabularx}{\linewidth}{
      @{}
      >{\raggedright\arraybackslash}p{0.12\linewidth}
      >{\raggedright\arraybackslash}p{0.34\linewidth}
      >{\raggedright\arraybackslash}p{0.12\linewidth}
      >{\raggedright\arraybackslash}X
      @{}
  }
    \toprule
    Symbol & Description & Symbol & Description \\
    \midrule

    \multicolumn{4}{@{}l}{\textit{Indices and dimensions}} \\
    \midrule
    $t=1,\ldots,T$ 
    & Decision-step index
    & $k=1,\ldots,K$ 
    & OD-pair index \\

    $l=1,\ldots,L$ 
    & Network-link index
    & $L^{\mathrm{obs}}$ 
    & Number of observed link-flow measurements \\

    \midrule
    \multicolumn{4}{@{}l}{\textit{Actions}} \\
    \midrule
    $a_{t,k}$ 
    & OD demand for OD pair $k$ at step $t$
    & $a_t$ 
    & Clipped OD demand vector applied to the DNL model \\

    $\widetilde{a}_t$ 
    & Raw Gaussian action sampled before clipping
    & $a^{\min},a^{\max}$ 
    & Lower and upper OD demand bounds \\

    $\mathcal{A}$ 
    & Feasible set of OD demand vectors
    & 
    & \\

    \midrule
    \multicolumn{4}{@{}l}{\textit{Observed and simulated link-flow variables}} \\
    \midrule
    $q_t^{*}$ 
    & Observed link-flow vector at step $t$
    & $e_t$ 
    & Simulated link-flow vector over all network links \\

    $H$ 
    & Selection matrix for extracting observed links
    & $q_t=He_t$ 
    & Simulated link-flow vector at observed links \\

    $c_l$ 
    & Capacity of link $l$ in vehicles per 15-minute interval
    & $c^{\mathrm{obs}}$ 
    & Capacity vector for observed link-flow measurements \\

    $\oslash$ 
    & Elementwise division
    & $d_l$
    & Input feature scale $c_l/15$ \\

    \midrule
    \multicolumn{4}{@{}l}{\textit{Network-state variables}} \\
    \midrule
    $n_{t,l}$ 
    & Link accumulation on link $l$ after step $t$
    & $v_{t,l}$ 
    & Relative speed on link $l$ after step $t$ \\

    $n_l^{\max}$ 
    & Maximum accumulation of link $l$
    & $\eta_t$ 
    & Normalised time index at step $t$ \\

    $s_t$ 
    & Observation summary provided to the policy
    & $\mathcal{S}$ 
    & Space of policy inputs \\

    $x_t$
    & Complete environment state
    &
    & \\

    \midrule
    \multicolumn{4}{@{}l}{\textit{Sequential estimator and policy objects}} \\
    \midrule
    $\pi$ 
    & Sequential estimator, or policy
    & $\Pi$ 
    & Set of feasible sequential estimators \\

    \midrule
    \multicolumn{4}{@{}l}{\textit{Link-flow propagation guidance}} \\
    \midrule
    $M_{t,\tau,k,l}$ 
    & Propagated volume on link $l$ at step $t$ generated by $a_{\tau,k}$
    & $\psi_{t,l}$ 
    & Reward sensitivity with respect to simulated flow \\

    $g_{\tau,k}^{\mathrm{LFPG}}$ 
    & LFPG signal for OD pair $k$ selected at step $\tau$
    &
    & \\

    \midrule
    \multicolumn{4}{@{}l}{\textit{LFPG normalisation and shaping terms}} \\
    \midrule
    $\mathcal{N}_{K}(\cdot)$ 
    & OD-wise normalisation across OD components
    & $\epsilon$ 
    & Small positive numerical constant \\

    $\alpha_A$ 
    & LFPG advantage-shaping strength
    & $\xi_{t,k}$ 
    & Standardised sampled-action residual \\

    $\kappa$ 
    & Clipping threshold for LFPG shaping
    & $S_{t,k}^{\mathrm{LFPG}}$ 
    & LFPG shaping component for OD pair $k$ \\

    \midrule
    \multicolumn{4}{@{}l}{\textit{PPO and temporal-advantage terms}} \\
    \midrule
    $\gamma$ 
    & Temporal weighting factor
    & $\lambda$ 
    & GAE trace-decay parameter \\

    $\overline{A}_{t}^{G}$ 
    & Normalised global advantage
    & $\widehat{R}_{t}$ 
    & Return target for critic training \\

    $\rho_{t,k}$ 
    & Component-wise PPO likelihood ratio
    & $\epsilon_{\mathrm{PPO}}$ 
    & PPO clipping parameter \\

    \bottomrule
  \end{tabularx}
\end{table}

\subsubsection{Dynamic OD matrix estimation problem}

For a target link-flow trajectory \(\{q_t^{*}\}_{t=1}^{T}\), a sequential estimator \(\pi\) selects demand vectors \(a_t\in\mathcal{A}\) using the information available at each step.

DNL maps the selected demand to network-link flows $e_t$, with $q_t=He_t$ selecting the observed links. Because vehicles propagate over time, $e_t$ depends on the demand history $a_1,\ldots,a_t$. The online DODE objective is

\begin{equation}
\label{eq:eq1}
\begin{aligned}
J(\pi)
&=
\mathbb{E}
\left[
\sum_{t=1}^{T}
\frac{1}{L^{\mathrm{obs}}}
\left\|
(q_t-q_t^{*}) \oslash c^{\mathrm{obs}}
\right\|_{2}^{2}
\right], \\
\pi^{*}
&\in
\operatorname*{arg\,min}_{\pi\in\Pi} J(\pi),
\end{aligned}
\end{equation}

where \(\Pi\) is the set of feasible sequential estimators. The expectation is over target trajectories, simulator and route-choice randomness, and any stochastic policy actions. The estimated dynamic OD matrix comprises the sequentially selected demand vectors.

\subsubsection{Sequential policy formulation}

Each episode begins by selecting a daily target trajectory and resetting DNL, inducing the initial-state distribution \(\rho_0\). The complete environment state \(x_t\) includes simulator queues and the process generating target observations. The policy receives only the observation summary \(s_t\), rather than the complete state:

\begin{equation}
\label{eq:eq2}
\begin{aligned}
s_t
&=
\left[
\eta_t,\,
q_t^{*}\oslash d^{\mathrm{obs}},\,
e_{t-1}\oslash d,\,
n_{t-1}\oslash n^{\max},\,
v_{t-1}
\right], \\
\eta_t
&=
\frac{t-1}{T-1},
\end{aligned}
\end{equation}

where \(c=(c_1,\ldots,c_L)\) is link capacity in vehicles per 15-minute interval, \(c^{\mathrm{obs}}=Hc\), and \(n^{\max}=(n_1^{\max},\ldots,n_L^{\max})\). Policy inputs are scaled by one-minute capacity \(d=c/15\), with \(d^{\mathrm{obs}}=Hd\). Rewards and LFPG sensitivities use the 15-minute capacity \(c\).

Counts \(q_t^*\) are available at the end of observation interval \(t\). The estimator uses these counts and the simulator state through \(t-1\) to calibrate demand for interval \(t\), then advances DNL through that interval. Thus, each estimate uses current and past observations, without access to future observations.

The initial simulator state is \(e_0=\mathbf{0}\), \(n_0=\mathbf{0}\), and \(v_0=\mathbf{1}\).

Applying $a_t$ to DNL produces link flows $e_t$, accumulations $n_t$, and relative speeds $v_t$. For $t<T$, the next policy input is \(s_{t+1}=[\eta_{t+1},\,q_{t+1}^{*}\oslash d^{\mathrm{obs}},\,e_t\oslash d,\,n_t\oslash n^{\max},\,v_t]\).

DNL and the process generating target observations define the transition kernel for the complete environment state:

\begin{equation}
\label{eq:eq3}
x_{t+1}\sim P(\cdot \mid x_t,a_t).
\end{equation}

The reward is defined as the negative normalised link-flow error,
\begin{equation}
\label{eq:eq4}
r(s_t,a_t,s_{t+1})
=
-\frac{1}{L^{\mathrm{obs}}}
\left\|
(q_t-q_t^{*})\oslash c^{\mathrm{obs}}
\right\|_{2}^{2}.
\end{equation}

Maximising cumulative reward minimises the objective in Eq.~\eqref{eq:eq1}. Rewards use observed links, while policy inputs include simulated conditions across the full network.

For a policy \(\pi_{\theta}(a_t\mid s_t)\) with parameters \(\theta\), the learning problem is

\begin{equation}
\label{eq:eq5}
\max_{\theta}
\mathbb{E}_{\rho_0,P,\pi_{\theta}}
\left[
\sum_{t=1}^{T}
r(s_t,a_t,s_{t+1})
\right].
\end{equation}

Equation~\eqref{eq:eq5} gives the undiscounted DODE objective. For policy training, GAE and LFPG use the temporal weight \(\gamma=0.99\).

\subsection{Policy Optimisation with Link-Flow Propagation Guidance}

\subsubsection{Extraction of link-flow propagation information}

LFPG uses the contributions of each OD demand component to link flows over a completed DNL rollout.

Vehicles generated by demand \(a_{\tau,k}\) form an OD-time group indexed by departure step \(\tau\) and OD pair \(k\). We record the volume from this group contributing to flow on link \(l\) at step \(t\) as

\begin{equation}
\label{eq:eq6}
\begin{gathered}
M_{t,\tau,k,l}\geq 0, \\
1\leq \tau \leq t \leq T,\quad
k=1,\ldots,K,\quad
l=1,\ldots,L.
\end{gathered}
\end{equation}

For \(t<\tau\), \(M_{t,\tau,k,l}=0\) because demand cannot contribute to flow before departure. The tensor

\begin{equation}
\label{eq:eq7}
M
=
\left\{M_{t,\tau,k,l}\right\}
\in
\mathbb{R}_{+}^{T\times T\times K\times L}
\end{equation}

records the realised propagation of all OD-time groups. For \(a_{\tau,k}>0\), \(M_{t,\tau,k,l}/a_{\tau,k}\) is the corresponding OD-to-link assignment fraction, which varies with departure time and network conditions \citep{ref14,ref16}.

\begin{figure}[htbp]
  \centering
  \includegraphics[width=1.0\linewidth]{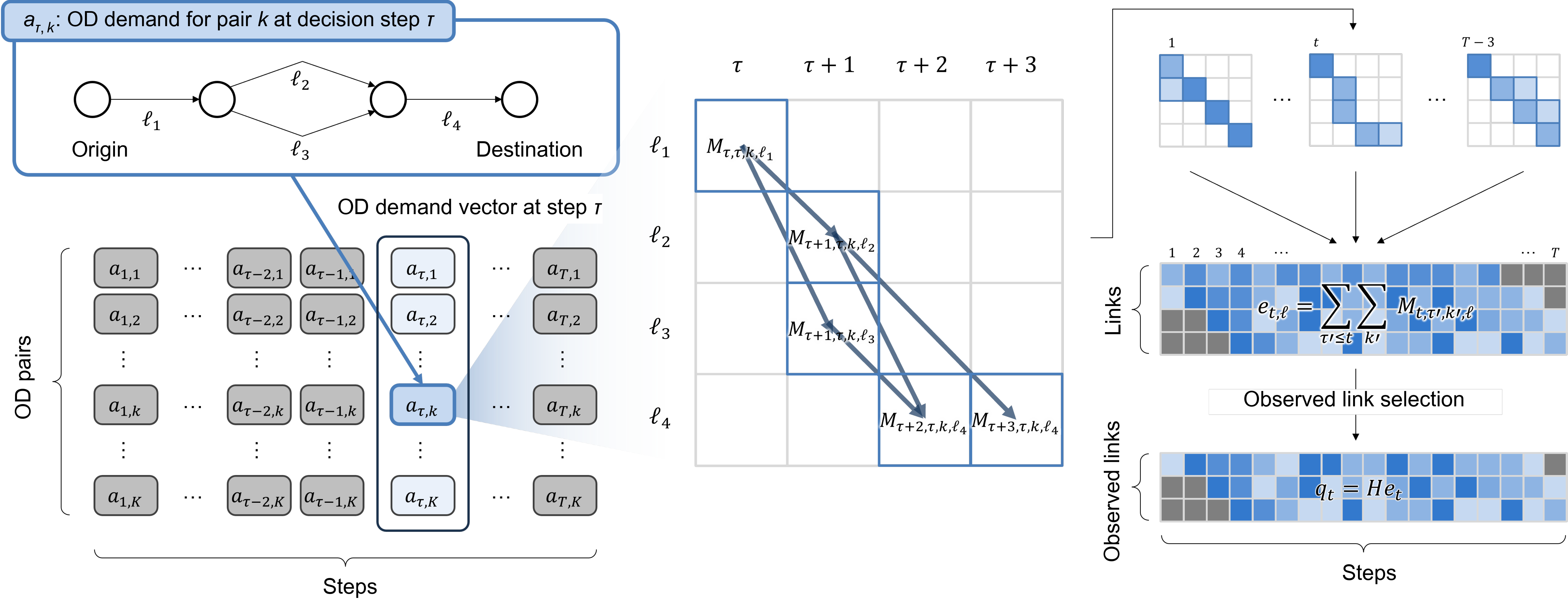}
  \caption{Extraction of link-flow propagation information from an OD-time demand component.}
  \label{fig2}
\end{figure}

As Fig.~\ref{fig2} illustrates, a single OD demand component can contribute to several downstream links over successive intervals.

Summing these contributions gives the simulated network-link flow and, through \(H\), the flow at observed links:

\begin{equation}
\label{eq:eq8}
e_{t,l}
=
\sum_{\tau=1}^{t}
\sum_{k=1}^{K}
M_{t,\tau,k,l},
\quad
l=1,\ldots,L,
\quad
q_t=He_t.
\end{equation}

To relate the recorded volumes to flow errors, we differentiate the reward with respect to simulated flow. Mapping this sensitivity to the full network through \(H^{\top}\) gives

\begin{equation}
\label{eq:eq9}
\psi_{t,l}
=
\left[
H^{\top}
\left(
\frac{2(q_t^{*}-q_t)}{L^{\mathrm{obs}}}
\oslash
(c^{\mathrm{obs}})^2
\right)
\right]_{l},
\quad
l=1,\ldots,L.
\end{equation}

Here, \((c^{\mathrm{obs}})^2\) denotes the component-wise square of \(c^{\mathrm{obs}}\). Network links not selected by \(H\) receive zero sensitivity.

A positive \(\psi_{t,l}\) indicates that simulated flow is below the target, and a negative value indicates that it exceeds the target.

For each OD-time component \((\tau,k)\), the LFPG signal is defined as

\begin{equation}
\label{eq:eq10}
g_{\tau,k}^{\mathrm{LFPG}}
=
\sum_{u=\tau}^{T}
\gamma^{u-\tau}
\sum_{l=1}^{L}
M_{u,\tau,k,l}\psi_{u,l},
\end{equation}

where \(\gamma\in(0,1]\) weights later contributions. The signal combines downstream reward sensitivities with the recorded volumes of demand component \(a_{\tau,k}\). Its sign suggests whether that demand should increase or decrease under the realised propagation pattern. Because changing demand can also alter routes and congestion, LFPG is a volume-weighted learning signal, not an exact DNL derivative.

\begin{figure}[htbp]
  \centering
  \includegraphics[width=1.0\linewidth]{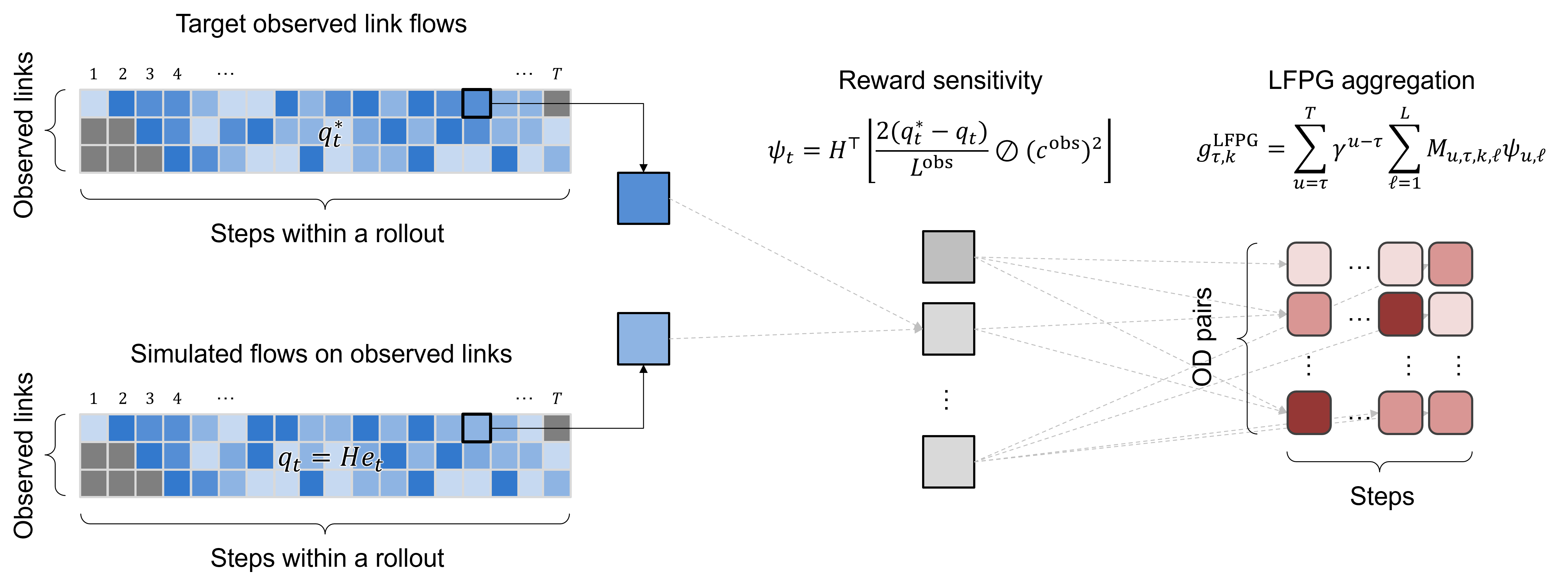}
  \caption{Construction of LFPG from propagated volumes and link-flow mismatch.}
  \label{fig3}
\end{figure}

Figure~\ref{fig3} summarises how propagation records and flow errors produce the OD-specific guidance in Eq.~\eqref{eq:eq10}.

Propagation recording and guidance extraction require storage and processing but no additional DNL evaluations. Both are used only during training. At deployment, the policy estimates demand from \(s_t\) alone.

\subsubsection{Proximal policy optimisation with LFPG}

PPO-Clip uses the likelihood ratio of the joint action in its clipped surrogate objective \citep{ref28}. Here, likelihood ratios and clipping are applied separately to each OD component so that each LFPG signal shapes the corresponding demand component without tying its clipping condition to likelihood changes in other components. The baseline denoted RL uses the same update with \(\alpha_A=0\), isolating the contribution of LFPG within the common policy architecture.

The policy is Gaussian with diagonal covariance, a state-dependent mean \(\mu_{\theta}(s_t)\in\mathbb{R}^{K}\), and a learned, state-independent standard deviation \(\sigma_{\theta}\in\mathbb{R}_{+}^{K}\). During training, raw actions are sampled as

\begin{equation}
\label{eq:eq11}
\widetilde{a}_t
\sim
\mathcal{N}
\left(
\mu_{\theta}(s_t),
\operatorname{diag}(\sigma_{\theta}^{2})
\right),
\end{equation}

DNL receives \(a_t=\operatorname{clip}(\widetilde a_t,0,30)\), measured in vehicles per 15-minute interval. The actor mean is bounded by \(\mu_\theta(s_t)=30\operatorname{sigmoid}(f_\theta(s_t))\). Likelihoods are evaluated at the raw sample, while deployment uses the deterministic mean. The exploration noise is independent across OD components conditional on the state. A shared neural network produces the state-dependent mean vector. The critic estimates \(V_\phi(s_t)\).

The global temporal learning signal is computed using generalised advantage estimation (GAE) \citep{ref15}. Let

\begin{equation}
\label{eq:eq12}
\delta_t
=
r(s_t,a_t,s_{t+1})
+
\gamma
\mathbf{1}_{\{t<T\}}
V_{\phi}(s_{t+1})
-
V_{\phi}(s_t),
\end{equation}

where \(\mathbf{1}_{\{t<T\}}\) is zero at the terminal step. The global advantage is

\begin{equation}
\label{eq:eq13}
\widehat{A}_{t}^{G}
=
\sum_{j=t}^{T}
(\gamma\lambda)^{j-t}\delta_j,
\end{equation}

where \(\lambda\) is the GAE trace-decay parameter. The value target is \(\widehat{R}_{t}=\widehat{A}_{t}^{G}+V_{\phi}(s_t)\). We denote the global advantage normalised over the rollout batch by \(\overline{A}_{t}^{G}\).

The global advantage is shared across OD components. LFPG supplies an additional signal for each component, with departure index \(\tau\) now written as \(t\) to match the action being updated. Let \(\theta_{\mathrm{old}}\) denote the policy parameters used to collect the rollout. We normalise LFPG by its mean absolute magnitude across OD components and combine it with the standardised sampled-action residual:

\begin{equation}
\label{eq:eq14}
\mathcal{N}_{K}
\left(
g_{t,k}^{\mathrm{LFPG}}
\right)
=
\operatorname{clip}_{\kappa}
\left(
\frac{
g_{t,k}^{\mathrm{LFPG}}
}{
\max\left(
\frac{1}{K}
\sum_{j=1}^{K}
\left|
g_{t,j}^{\mathrm{LFPG}}
\right|
,\epsilon\right)
}
\right),
\end{equation}

\begin{equation}
\label{eq:eq15}
\xi_{t,k}
=
\operatorname{clip}_{\kappa}
\left(
\frac{
\widetilde{a}_{t,k}
-
\mu_{\theta_{\mathrm{old}},k}(s_t)
}{
\sigma_{\theta_{\mathrm{old}},k}
}
\right),
\end{equation}

\begin{equation}
\label{eq:eq16}
S_{t,k}^{\mathrm{LFPG}}
=
\alpha_{A}
\operatorname{clip}_{\kappa}
\left(
\mathcal{N}_{K}
\left(
g_{t,k}^{\mathrm{LFPG}}
\right)
\xi_{t,k}
\right),
\end{equation}

where \(\operatorname{clip}_{\kappa}(z)=\min(\kappa,\max(-\kappa,z))\), \(\epsilon=10^{-6}\) prevents division by zero, and \(\alpha_A\geq0\) controls the shaping strength. The product in Eq.~\eqref{eq:eq16} favours sampled demand perturbations aligned with the guidance direction.

\begin{figure}[htbp]
  \centering
  \includegraphics[width=1.0\linewidth]{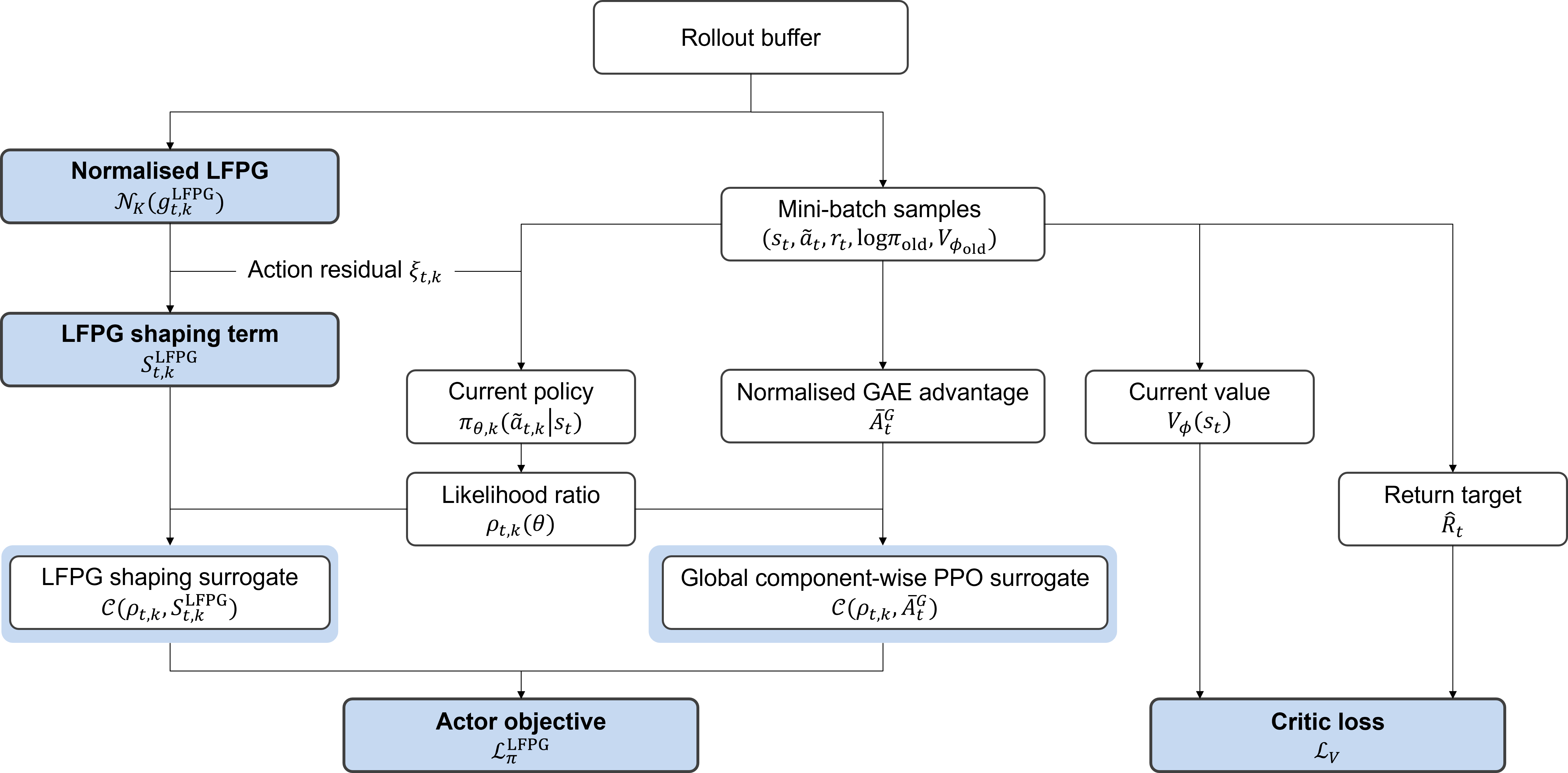}
  \caption{LFPG-RL policy update with global PPO advantage and OD-specific LFPG shaping.}
  \label{fig4}
\end{figure}

Figure~\ref{fig4} shows how global and OD-specific signals enter the policy update. For the marginal Gaussian density \(\pi_{\theta,k}\) of OD component \(k\), the component-wise likelihood ratio is

\begin{equation}
\label{eq:eq17}
\rho_{t,k}(\theta)
=
\frac{
\pi_{\theta,k}
\left(
\widetilde{a}_{t,k}\mid s_t
\right)
}{
\pi_{\theta_{\mathrm{old}},k}
\left(
\widetilde{a}_{t,k}\mid s_t
\right)
}.
\end{equation}

The LFPG-RL actor objective is then

\begin{equation}
\label{eq:eq18}
\begin{aligned}
\mathcal{C}(\rho,x)
&=
\min
\left[
\rho x,\,
\operatorname{clip}
\left(
\rho,
1-\epsilon_{\mathrm{PPO}},
1+\epsilon_{\mathrm{PPO}}
\right)x
\right], \\
\mathcal{L}_{\pi}^{\mathrm{LFPG}}(\theta)
&=
\frac{1}{T}
\sum_{t=1}^{T}
\sum_{k=1}^{K}
\left[
\mathcal{C}
\left(
\rho_{t,k}(\theta),
\overline{A}_{t}^{G}
\right)
+
\mathcal{C}
\left(
\rho_{t,k}(\theta),
S_{t,k}^{\mathrm{LFPG}}
\right)
\right],
\end{aligned}
\end{equation}

where \(\epsilon_{\mathrm{PPO}}\) controls likelihood-ratio clipping. Clipping is applied separately to the normalised global advantage and LFPG shaping term. The critic minimises the squared error between \(V_{\phi}(s_t)\) and \(\widehat{R}_{t}\), and the training loss includes entropy regularisation.

\subsection{Experimental Settings}

\subsubsection{Dynamic network loading model}

The DNL model is a discrete-time link transmission model (LTM) with sending and receiving constraints, finite link storage, and backward-wave travel times \citep{ref29,ref30,ref31}. Estimation uses 15-minute intervals aggregated from one-minute internal steps. The demand for each OD pair is distributed uniformly over the fifteen loading steps. At most four simple candidate paths per OD pair are generated from free-flow costs and held fixed. A stochastic logit model updates route probabilities from current network costs at each internal minute, using a cost coefficient of 0.5 and a sampling unit of one vehicle. Each daily episode starts with an empty network. Future OD demand is not specified in this sequential estimation setting. We therefore use sequential network loading with stochastic route choice based on current network costs, rather than solving a full-horizon dynamic user equilibrium.

\subsubsection{Target network and data}

\begin{figure}[htbp]
  \centering
  \includegraphics[width=1.0\linewidth]{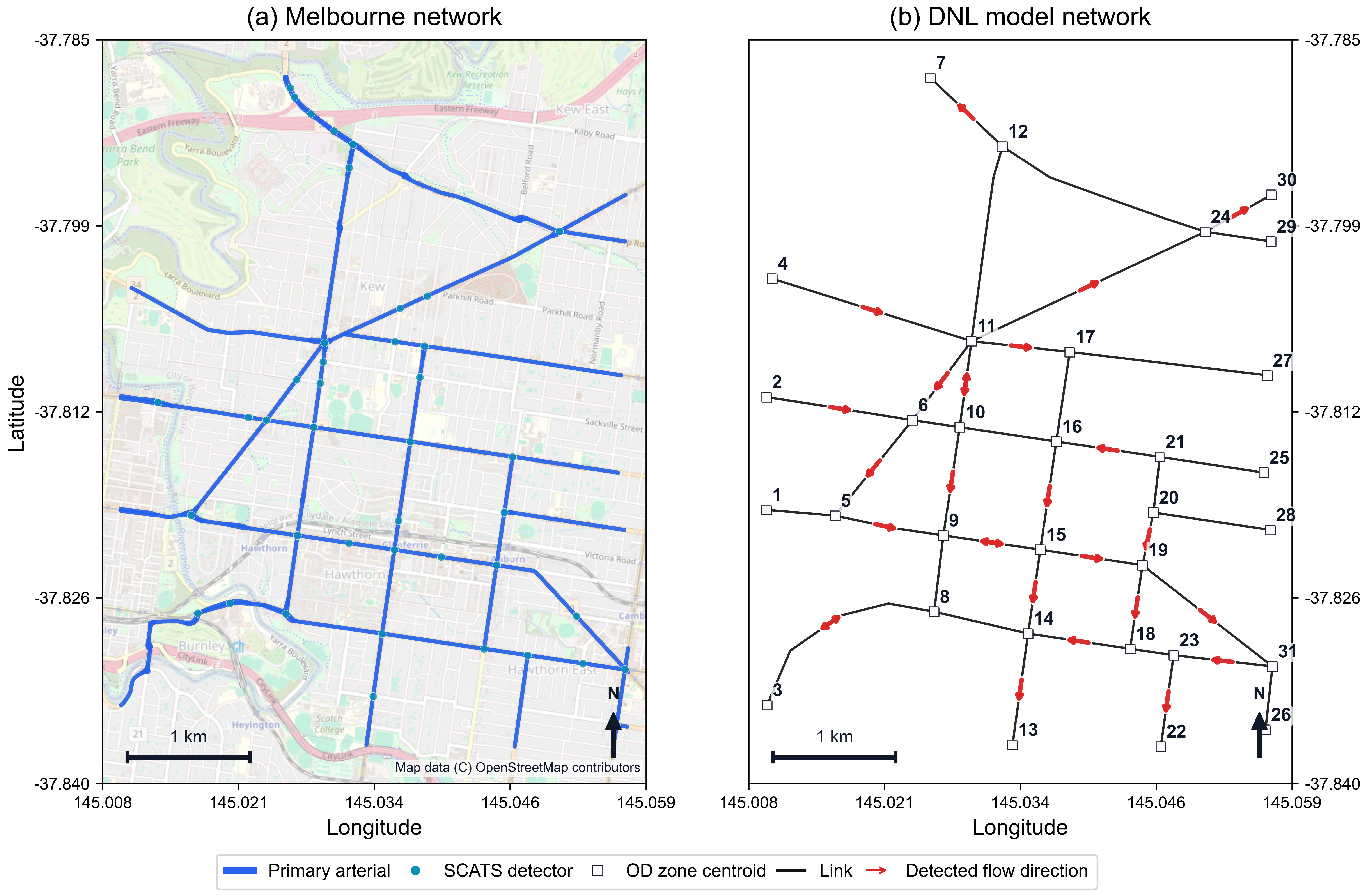}
  \caption{(a) Melbourne arterial network and (b) corresponding DNL model network.}
  \label{fig5}
\end{figure}

The case study uses a Melbourne arterial network and SCATS detector data (Fig.~\ref{fig5}). The DNL network has 31 OD zone centroids and 78 directed links, giving 930 OD components per action. Detectors are mapped to directed links, and one detector per observed link is selected based on data completeness. The selected detector count series are used directly as calibration targets. Link parameters follow Victorian strategic transport modelling values for urban arterials: 900 vehicles per hour per lane, a posted-speed factor of 0.75, and Akcelik \(J=0.8\) \citep{ref32}.

The dataset comprises 250 weekday link-flow trajectories retained after quality screening of SCATS records. The earliest 190 dates are used for training (1 April 2025--15 January 2026), the next 30 for validation (16 January--27 February 2026), and the latest 30 for testing (2 March--22 April 2026). Representative detectors for the 27 observed links are selected using training dates and then fixed for validation and testing. Each episode covers 04:00--10:00 in 24 intervals.

Because ground-truth OD matrices are unavailable, performance is evaluated by the agreement between simulated and observed link flows.

\subsubsection{Baselines and hyperparameters}

The comparison includes weighted simultaneous perturbation stochastic approximation (W-SPSA), the ensemble Kalman filter (EnKF), reinforcement learning without LFPG (RL), gradient descent with LFPG (LFPG-GD), EnKF with LFPG (LFPG-EnKF), and LFPG-RL. The RL baseline shares the actor, critic, optimiser, training budget, and checkpoint selection procedure of LFPG-RL. LFPG-GD uses realised assignment fractions to construct its search direction, while LFPG-EnKF refines posterior demand candidates using propagation guidance. These comparisons examine whether propagation information is also useful in conventional methods and distinguish its use in online search from its use in policy learning.

Every OD component lies in \([0,30]\) vehicles per 15-minute interval. The shared encoder is Linear(256)--Tanh--Linear(256), followed by separate mean and value heads. The learning rate is \(3\times10^{-4}\), \(\gamma=0.99\), GAE \(\lambda=0.95\), PPO clipping parameter 0.1, and batch size 96. Each update uses up to four epochs, with further optimiser steps stopped if the mean marginal approximate Kullback--Leibler (KL) divergence exceeds 0.045. The entropy coefficient is \(2\times10^{-4}\), value coefficient 0.5, and gradient-norm limit 0.5. The actor objective sums across OD coordinates and averages across the batch, while entropy averages across both.

The nominal initial policy mean is set to \StudyInitialMean{} vehicles and the initial standard deviation to 0.35. Each policy is trained for 16,384 daily episodes using four environments, with three independent runs. Checkpoints are evaluated every 1,024 episodes on the 30 validation dates, and the checkpoint with the lowest validation RMSE is selected. LFPG uses \(\alpha_A=1.35\) and \(\kappa=1.4\). Hyperparameter comparisons are reported in Appendix~\ref{app:hyperparameters}.

LFPG-GD uses zero initialisation and an initial step size of 0.8, while W-SPSA uses previous-step initialisation with \(a=0.5\). Both EnKF variants use 32 members, with process-noise scales of 0.01 for EnKF and 0.03 for LFPG-EnKF. Iterative methods receive a nominal 30 s search budget per interval, with at most 64 GD/SPSA iterations or eight filter cycles. Their parameter selection is described in Appendix~\ref{app:hyperparameters}.

The EnKF methods update an ensemble of current OD demand vectors and carry the centred ensemble deviations forward to the next interval. The evaluated candidate with the lowest capacity-normalised mean squared link-flow error is retained. W-SPSA uses current-interval OD-to-link weights from propagation records and an initial perturbation magnitude of 1.2 vehicles.

\section{Results}

\subsection{Link-flow estimation accuracy}

We evaluate root mean squared error (RMSE), mean absolute error (MAE), mean absolute percentage error (MAPE), and Pearson correlation on the 30 test dates, pooling 19,440 link--interval observations per run. Table~\ref{tab:accuracy} reports the mean and sample standard deviation across three runs. MAPE excludes the 974 observations with zero counts in each run.

\begin{table}[htbp]
\centering
\caption{Test link-flow estimation accuracy over three runs. RMSE and MAE are in vehicles per 15-minute interval, and MAPE is a percentage.}
\label{tab:accuracy}
\begin{tabular}{lrrrr}
\toprule
Method & RMSE & MAE & MAPE (\%) & Pearson $r$ \\
\midrule
W-SPSA & $33.81\pm0.03$ & $21.16\pm0.06$ & $125.04\pm0.37$ & $0.714\pm0.001$ \\
EnKF & $38.96\pm0.23$ & $25.73\pm0.09$ & $139.74\pm0.96$ & $0.671\pm0.002$ \\
RL & $48.58\pm1.92$ & $39.84\pm1.44$ & $524.76\pm47.89$ & $0.075\pm0.039$ \\
LFPG-GD & $12.38\pm0.12$ & $7.38\pm0.06$ & $42.42\pm0.04$ & $0.961\pm0.001$ \\
LFPG-EnKF & $12.88\pm0.44$ & $8.35\pm0.13$ & $46.17\pm0.57$ & $0.957\pm0.003$ \\
{\bfseries\boldmath LFPG-RL} & {\bfseries\boldmath $7.41\pm0.08$} & {\bfseries\boldmath $4.74\pm0.01$} & {\bfseries\boldmath $31.24\pm0.58$} & {\bfseries\boldmath $0.987\pm0.000$} \\
\bottomrule
\end{tabular}
\end{table}

LFPG-RL achieves the lowest errors and highest correlation among the six methods. Its RMSE of 7.41 vehicles per interval is 40.1\% lower than that of LFPG-GD, the strongest baseline, and 84.7\% lower than that of RL. The paired RMSE difference (LFPG-RL minus LFPG-GD) is -4.97, with a 95\% confidence interval of [-6.01, -3.85], estimated from 5,000 paired bootstrap resamples of runs and five-date circular blocks. LFPG-RL also has lower daily RMSE in 87 of 90 paired date--run comparisons. Figures~\ref{fig7} and \ref{fig8} show the agreement with observations and the distribution of daily errors.

\subsection{Policy learning}

\begin{figure}[htbp]
\centering
\includegraphics[width=\linewidth]{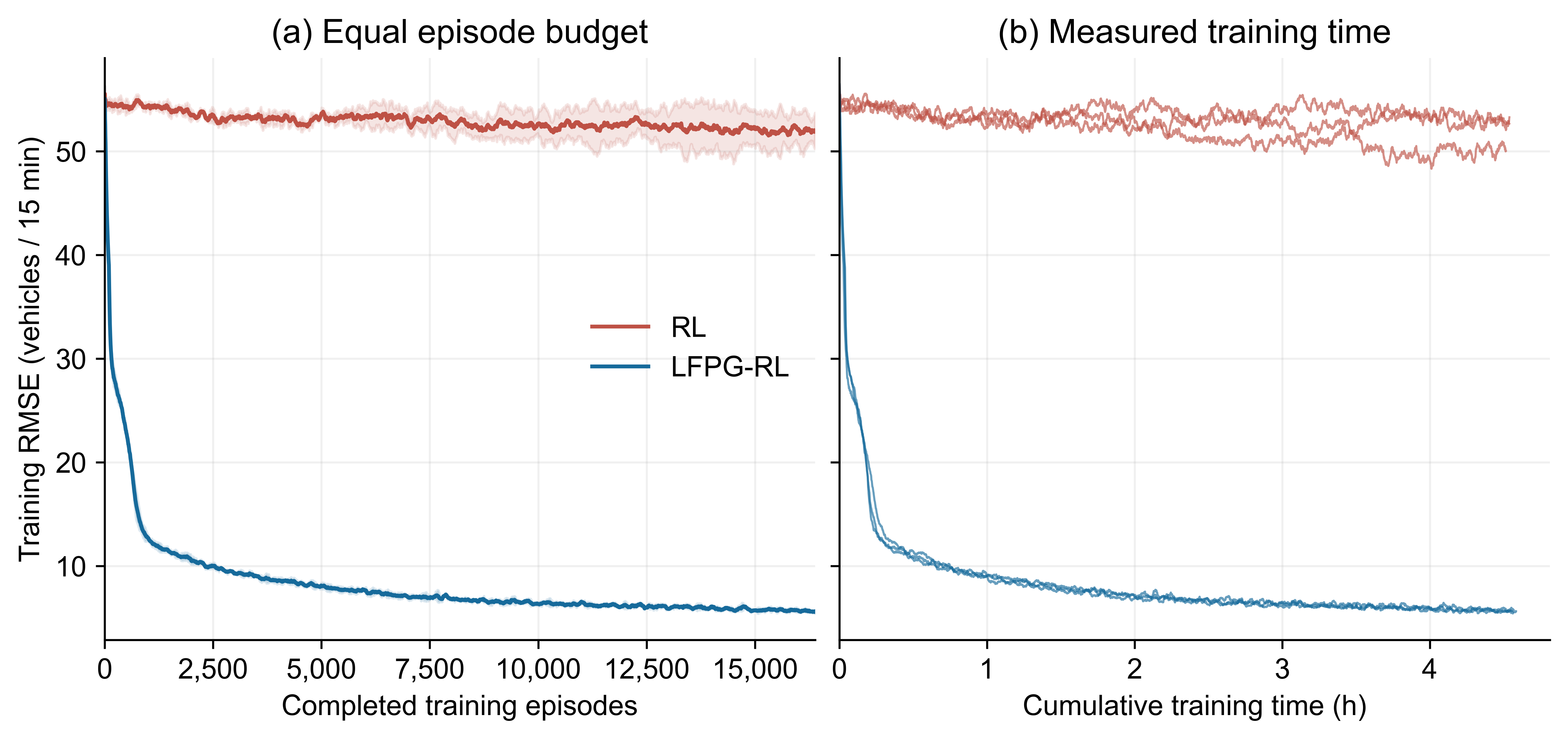}
\caption{Training RMSE over trailing windows of up to 100 episodes, starting with the first rollout batch. (a) Mean \(\pm\) one sample standard deviation across three runs. (b) Individual runs against cumulative training time.}
\label{fig6}
\end{figure}

\begin{figure}[htbp]
\centering
\includegraphics[width=\linewidth]{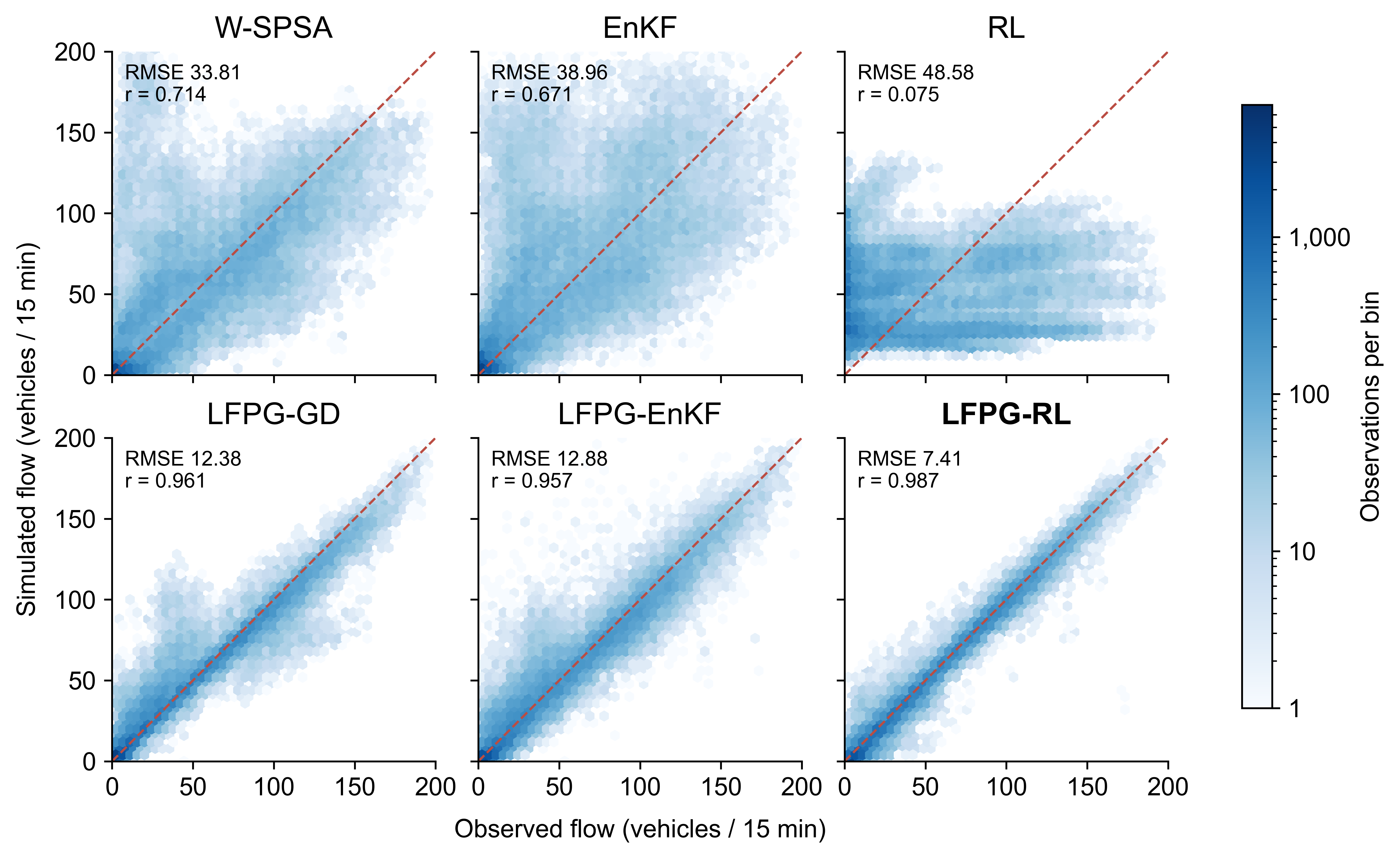}
\caption{Observed and simulated link flows across three runs and 30 test dates. Hexagonal-bin shading shows counts on a common logarithmic colour scale. Axes show the 0--200 vehicle range. Metrics include all observations. Annotations report mean per-run RMSE and correlation, as in Table~\ref{tab:accuracy}.}
\label{fig7}
\end{figure}

\begin{figure}[htbp]
\centering
\includegraphics[width=\linewidth]{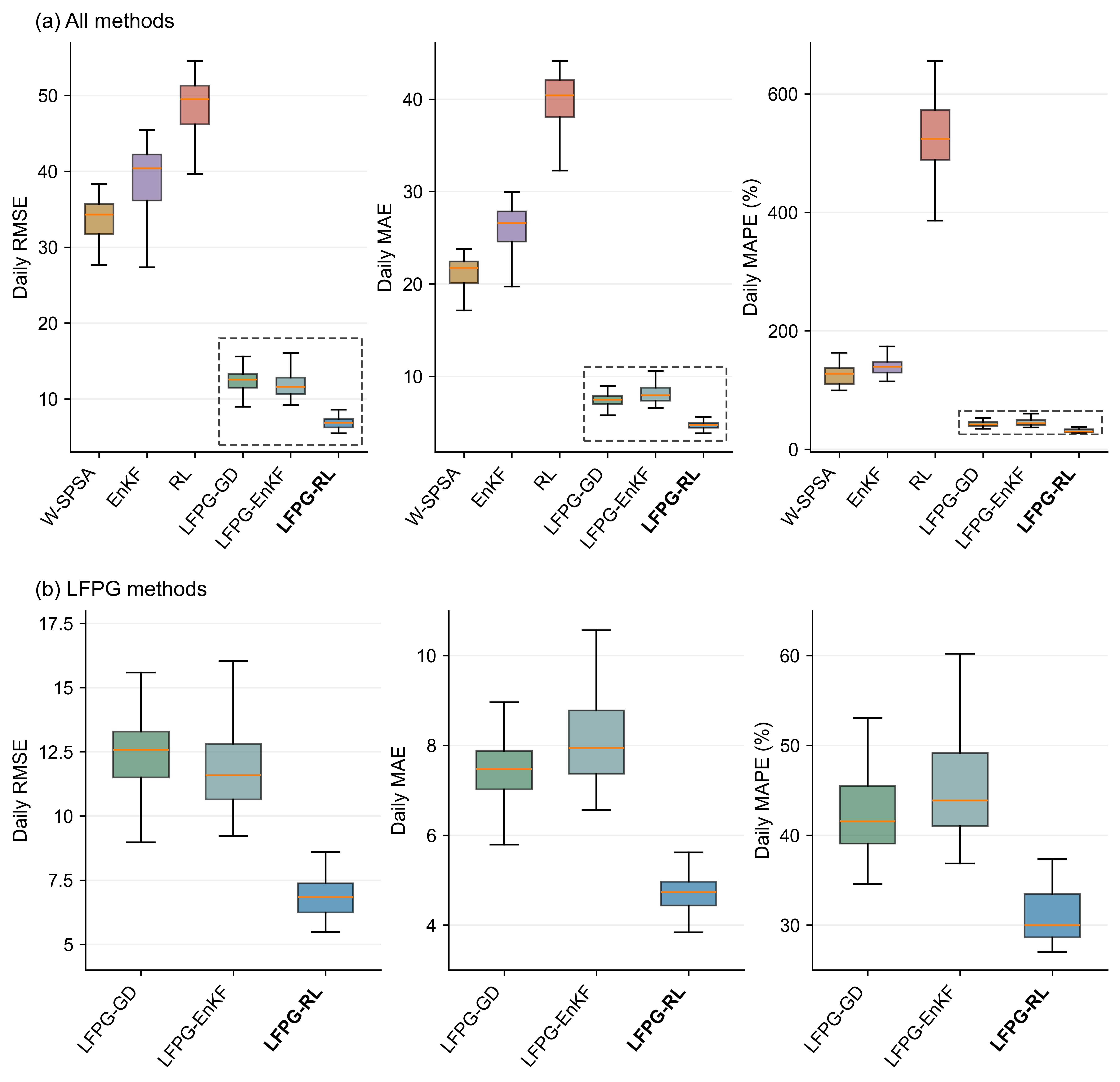}
\caption{Daily errors across 30 test dates and three runs. (a) All six methods, with dashed rectangles marking the LFPG methods and the ranges enlarged in (b). (b) Detailed comparisons of LFPG-GD, LFPG-EnKF, and LFPG-RL. Boxes show medians and interquartile ranges, with whiskers spanning values within 1.5 interquartile ranges of each box. Points beyond the whiskers are hidden but retained in numerical summaries.}
\label{fig8}
\end{figure}

LFPG-RL reaches lower training errors than RL in all three runs under the same 16,384-episode budget (Fig.~\ref{fig6}). Within each paired run, the methods start from the same policy and have identical errors before the first update. The subsequent difference in training errors indicates the benefit of LFPG under the shared training configuration. The lowest validation RMSE occurs between 10,240 and 16,384 episodes across the six runs.

\subsection{Computational performance}

\begin{table}[htbp]
\centering
\caption{Processing time per observation interval, including demand estimation and the DNL transition.}
\label{tab:online_cost}
\begin{tabular}{lrrr}
\toprule
Method & Median (s) & 95th percentile (s) & Maximum (s) \\
\midrule
W-SPSA & 26.938 & 30.720 & 31.279 \\
EnKF & 30.141 & 30.464 & 30.738 \\
{\bfseries\boldmath RL} & {\bfseries\boldmath 0.032} & {\bfseries\boldmath 0.089} & {\bfseries\boldmath 0.977} \\
LFPG-GD & 29.101 & 29.296 & 29.607 \\
LFPG-EnKF & 30.113 & 30.433 & 36.396 \\
{\bfseries\boldmath LFPG-RL} & {\bfseries\boldmath 0.012} & {\bfseries\boldmath 0.053} & {\bfseries\boldmath 0.673} \\
\bottomrule
\end{tabular}
\end{table}

LFPG-RL processes an interval in a median of 0.012 s, including policy evaluation and one DNL transition. Median policy evaluation takes 0.458 ms. Under the specified search budget, the iterative methods have median processing times of approximately 27--30 s per interval (Table~\ref{tab:online_cost}). Measurements use an AMD Ryzen Threadripper PRO 5975WX with 128 GB RAM. Offline training takes 4.57\(\pm\)0.02 hours per LFPG-RL run and 4.53\(\pm\)0.01 hours per RL run.

\subsection{Spatial and temporal error patterns}

\begin{figure}[htbp]
\centering
\includegraphics[width=\linewidth]{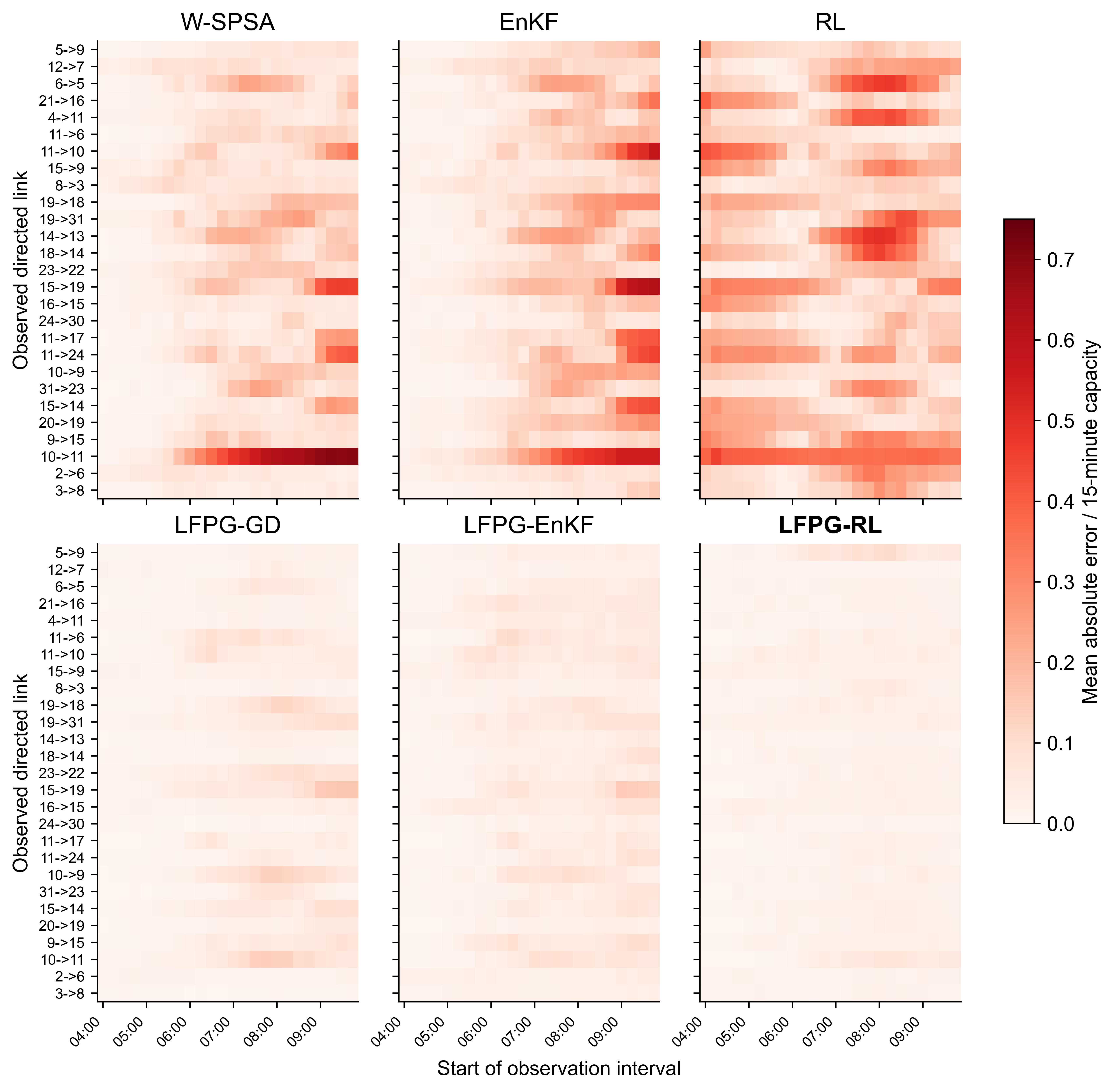}
\caption{Absolute link-flow error divided by 15-minute capacity, averaged over 30 test dates and three runs. All six methods share a colour scale. Rows are observed directed links and columns are observation intervals.}
\label{fig9}
\end{figure}

Figure~\ref{fig9} compares the distribution of errors across observed links and time intervals for all six methods. LFPG-RL has 24 of 648 link--interval cells with mean absolute error above 5\% of 15-minute capacity, compared with 119 for LFPG-GD. LFPG-RL thus has substantially fewer cells with large errors. The comparison with RL indicates that propagation guidance helps the common policy architecture learn from varying target flows.

\section{Conclusion}
Online DODE requires sequential demand updates that reproduce observed traffic conditions within short decision intervals. Repeated simulator evaluations can make this difficult, while RL must learn a common estimation rule across varying target flow trajectories. This study proposed LFPG-RL to address these challenges by combining downstream flow errors with simulated vehicle propagation records, providing OD-specific guidance during policy learning. The trained policy estimates current demand from available observations and the simulated network state in a single evaluation.

In the Melbourne arterial network, LFPG-RL achieved a mean test RMSE of 7.41 vehicles per 15-minute interval, MAPE of 31.24\%, and Pearson correlation of 0.987. It reduced RMSE by 40.1\% relative to LFPG-GD, the strongest baseline, and by 84.7\% relative to RL without propagation guidance. Median online processing took 0.012 s per interval, including the DNL transition, compared with approximately 27--30 s for iterative methods.

This study introduces propagation-based guidance that links aggregate link-flow errors to individual OD demand components during policy learning. LFPG uses realised propagation information to guide policy updates. Policy training is performed offline, enabling rapid demand updates during online application. The resulting demand updates keep traffic simulations consistent with observed conditions and support the evaluation of traffic management strategies.

Future work will examine larger networks and heavier congestion, where longer propagation delays may alter the information needed by the policy. Further evaluation under sparse or noisy observations and changing traffic regimes would clarify how training data and detector coverage affect estimation. Incorporating additional observations, such as travel times, may also help distinguish demand patterns that produce similar link flows.

\section*{Acknowledgements}
The authors used GPT-6 Astra (OpenAI) for language editing and coding assistance. The authors reviewed and revised the text and independently verified the code and results.

\section*{Author Contributions}
Min and Kim contributed to conceptualisation, methodology, and manuscript writing and revision. Min performed data curation and formal analysis. Both authors reviewed the results and approved the final manuscript.

\section*{Code Availability}
The source code and the training, validation and test datasets are available at \url{https://github.com/dgmin-kr/dode-rl-online/}.

\section*{Declaration of Conflicting Interests}
The authors declared no potential conflicts of interest with respect to the research, authorship, and/or publication of this article.

\section*{Funding}
This work was supported by the Korea Institute of Police Technology (No. 092021C28S02000) and the National Research Foundation of Korea (No. 00409860).


\bibliographystyle{tTMB}
\bibliography{references}

\appendices
\section{Hyperparameter comparisons}
\label{app:hyperparameters}

Input and reward scaling were compared using one run per configuration, with an initial policy mean of 0.1 vehicles and 4,096 training episodes. Validation RMSE was evaluated on six dates (Table~\ref{tab:normalization}). One-minute capacity for inputs and 15-minute capacity for rewards were retained for both policies, giving the lowest LFPG-RL validation error among the tested settings.

\begin{table}[htbp]
\centering
\caption{Validation RMSE under alternative input and reward scales.}
\label{tab:normalization}
\begin{tabular}{lrr}
\toprule
Input / reward capacity & RL & LFPG-RL \\
\midrule
1 min / 1 min & 52.01 & 12.04 \\
15 min / 15 min & 58.88 & 18.60 \\
1 min / 15 min & 58.57 & 10.14 \\
\bottomrule
\end{tabular}
\end{table}

Nominal initial means of 0, 0.1, and 1 vehicle were compared after 1,024 episodes on the same six validation dates. A common initial mean of 1 vehicle was selected by the average validation RMSE of the two policies (Table~\ref{tab:initialisation}). The nominal zero is represented by a small positive mean because the actor uses a sigmoid output.

\begin{table}[htbp]
\centering
\caption{Validation RMSE for nominal initial policy means.}
\label{tab:initialisation}
\begin{tabular}{rrrr}
\toprule
Initial mean & RL & LFPG-RL & Average \\
\midrule
0.0 & 63.48 & 18.45 & 40.96 \\
0.1 & 58.48 & 15.95 & 37.22 \\
1.0 & 51.38 & 15.64 & 33.51 \\
\bottomrule
\end{tabular}
\end{table}

For the iterative methods, parameters were selected by RMSE on three validation dates. For LFPG-GD, zero and previous-step initialisation were compared with initial step sizes of 0.4 and 0.8. W-SPSA used the same initialisation choices with \(a\in\{0.5,2.0\}\). For each EnKF variant, ensemble sizes of 12 and 32 were evaluated with process-noise scales of 0.01, 0.03, 0.1, 0.3, and 0.6.

The LFPG weight was also examined with one run per setting under the main training budget and checkpoint selection procedure (Table~\ref{tab:sensitivity}). The zero-weight and 1.35-weight settings reuse one run from the main comparison. Weights of 0.675--2.7 give RMSEs of 7.19--7.89 vehicles per interval, indicating that the improvement over RL persists across the tested non-zero weights.

\begin{table}[htbp]
\centering
\caption{Test accuracy for different LFPG weights in a single-run sensitivity analysis.}
\label{tab:sensitivity}
\begin{tabular}{rrr}
\toprule
$\alpha_A$ & RMSE & MAE \\
\midrule
0.000 & 50.19 & 40.70 \\
0.675 & 7.89 & 5.09 \\
1.350 & 7.33 & 4.75 \\
2.700 & 7.19 & 4.69 \\
\bottomrule
\end{tabular}
\end{table}


\end{document}